\documentclass[conference]{IEEEtran}

\usepackage{times}
\usepackage[numbers]{natbib}
\usepackage[T1]{fontenc}
\usepackage[utf8]{inputenc}
\usepackage{amsmath,amssymb}
\usepackage{graphicx}
\usepackage{url}
\usepackage{booktabs}
\usepackage{array}
\usepackage{float}
\usepackage{placeins}
\usepackage{needspace}
\usepackage{algorithm}
\usepackage{algpseudocode}
\usepackage{etoolbox}
\usepackage{caption}
\usepackage{balance}
\usepackage[svgnames]{xcolor}
\usepackage{tikz}
\usepackage{fancyhdr}
\usetikzlibrary{
  positioning,arrows.meta,calc,shapes.geometric,fit,backgrounds
}
\usepackage[bookmarks=true]{hyperref}

\newcommand{\arxivciteas}{%
  Cite as: Z.~Meng and F.~Cruz,
  ``Initiation Safety: A Missing Dimension in Generalist-Robot Safety,''
  in \emph{Rethinking What It Means to be Safe for Generalist Robots Workshop}
  at Robotics: Science and Systems (RSS)~2026.%
}
\fancypagestyle{arxivcite}{%
  \fancyhf{}%
  \fancyhead[L]{%
    \parbox[t]{\textwidth}{%
      \raggedright\fontsize{7}{8.5}\selectfont
      \arxivciteas
    }%
  }%
}
\definecolor{MenTeal}{RGB}{18,120,115}
\definecolor{MenTealLight}{RGB}{220,242,240}
\definecolor{MenCoral}{RGB}{210,85,70}
\definecolor{MenCoralLight}{RGB}{255,236,232}
\definecolor{MenSlate}{RGB}{72,82,96}
\definecolor{MenSlateLight}{RGB}{244,246,249}
\definecolor{MenGold}{RGB}{196,140,40}

\makeatletter
\apptocmd{\thebibliography}{\small}{}{}
\makeatother

\title{Initiation Safety: A Missing Dimension in\\
Generalist-Robot Safety}

\author{
\authorblockN{Zhijin Meng and Francisco Cruz}
\authorblockA{School of Computer Science and Engineering\\
University of New South Wales\\
Email: zhijin.meng@unsw.edu.au, f.cruz@unsw.edu.au}
}

\begin{document}

\maketitle
\thispagestyle{arxivcite}

\begin{abstract}
Safety for generalist robots is usually discussed in terms of motion or
dialogue. We argue a third question is missing: should the robot take
its first hard-to-undo social action at all, such as a greeting, an uninvited
grasp, or stepping into someone's space? We call this \emph{initiation
authorization}. Current frameworks rarely treat it as a separate safety
layer. Today's stacks often skip this step: a high engagement score or
a confident VLA rollout is treated as permission to act. But seeing a
person is not the same as having their consent to be addressed. We
frame initiation authorization within generalist-robot safety and
contrast it with post-plan VLA guardrails, implementing \textsc{pas}
(probe--authorize--speak) on a doorway humanoid, comparing it with
direct-init on logged traces, and proposing a three-condition user study,
with open questions on metrics, governance, and where initiation ends
and foundation-model generation begins.

\end{abstract}

\section{Introduction}
\label{sec:intro}
Generalist-robot safety must include \emph{initiation
authorization}: deciding whether a robot may take its first
hard-to-undo social action when the situation is still unclear. An
engagement score measures how receptive someone seems. It does not
answer whether the robot should act when that signal is ambiguous
\citep{rich2010,broek2024proactive,shi2011spatial}.

Deployed stacks often distinguish two safety layers (physical and
social-interaction) but treat initiation as part of the latter rather
than as a separate decision. \emph{Physical safety} limits motion
(collisions, force, joint limits \citep{zhang2025safevla}).
\emph{Social-interaction safety} governs how the robot behaves once
interaction has already started, including recovery when a first act
misfires \citep{akalin2021,tian2021socialerrors,harland2025aiapology}.
We argue for a third layer that this picture largely omits:
\emph{initiation authorization}, whether the robot may start at all.
The harm
often looks like an
ordinary first move: a mistimed greeting that interrupts a call
\citep{edirisinghe2024shopworker}, or an end-to-end VLA that reaches
for an object before the person has agreed~\citep{openvla-2024}. A jailbroken planner can
authorize a harmful first move as readily as a harmful first word
\citep{robey2024jailbreaking}.

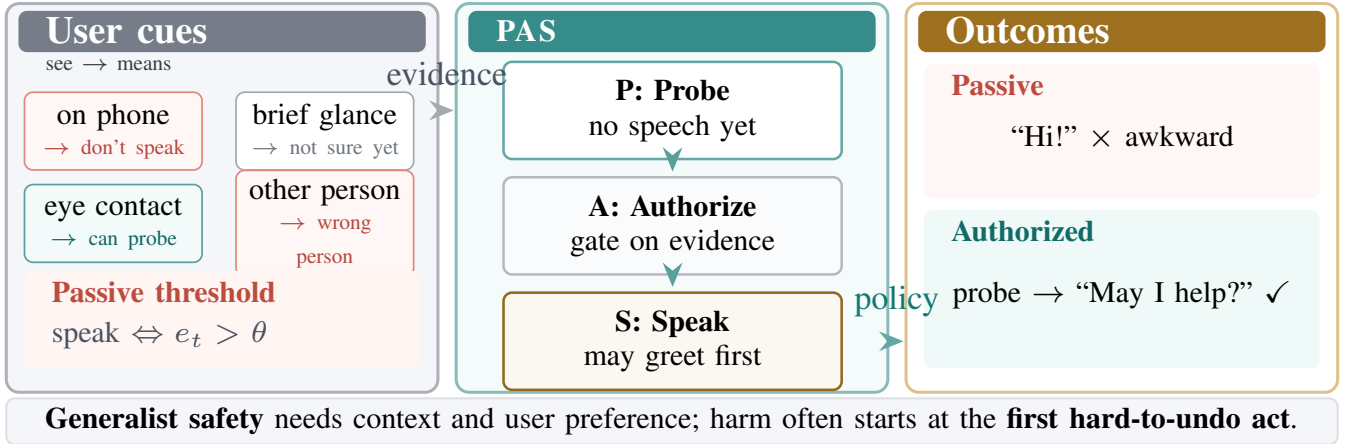
\begin{figure*}[t]
\centering
\resizebox{0.98\textwidth}{!}{%
\begin{tikzpicture}[
  font=\footnotesize,
  >=Stealth,
  arr/.style={-{Stealth[length=2mm,width=1.6mm]}, line width=0.9pt},
  panel/.style={rounded corners=3pt, line width=0.7pt},
  hdr/.style={font=\small\bfseries, text=white},
  card/.style={rounded corners=2pt, align=center, font=\scriptsize,
    inner sep=2pt, text width=1.32cm, minimum height=0.64cm},
  mod/.style={rounded corners=2pt, align=center, font=\scriptsize,
    inner sep=4pt, line width=0.6pt, text width=2.5cm, minimum height=0.80cm},
]
\def\colW{3.55}
\def\yBot{0.35}
\def\yTop{3.55}
\def\yHdrBot{3.15}
\def\xA{0.05}
\def\xB{3.75}
\def\xC{7.45}
\def\cueXL{0.93}      
\def\cueXR{2.67}      
\def\cueYtop{2.50}    
\def\cueYbot{1.74}    
\def\pasX{5.53}        
\def\pasGap{0.95}      
\def\pasYmid{1.72}    

\fill[MenSlate, rounded corners=2pt] (0,3.72) rectangle (11.05,4.02);
\node[font=\small\bfseries, text=white] at (5.52,3.87)
  {Initiation authorization under uncertainty};

\fill[MenSlateLight, panel] (\xA,\yBot) rectangle (\xA+\colW,\yTop);
\draw[MenSlate!45, panel] (\xA,\yBot) rectangle (\xA+\colW,\yTop);
\fill[MenSlate!75, rounded corners=2pt] (\xA+0.1,\yHdrBot) rectangle (\xA+\colW-0.1,\yTop-0.06);
\node[hdr, anchor=west] at (\xA+0.2,\yTop-0.22) {User cues};
\node[font=\tiny, text=MenSlate!75!black, anchor=west] at (\xA+0.2,3.04)
  {see $\rightarrow$ means};
\node[card, draw=MenCoral!70, fill=MenCoralLight!40] at (\cueXL,\cueYtop)
  {on phone\\[-1pt]\textcolor{MenCoral!85!black}{\tiny$\rightarrow$ don't speak}};
\node[card, draw=MenSlate!50, fill=white] at (\cueXR,\cueYtop)
  {brief glance\\[-1pt]\textcolor{MenSlate!80}{\tiny$\rightarrow$ not sure yet}};
\node[card, draw=MenTeal!70, fill=MenTealLight!35] at (\cueXL,\cueYbot)
  {eye contact\\[-1pt]\textcolor{MenTeal!85!black}{\tiny$\rightarrow$ can probe}};
\node[card, draw=MenCoral!70, fill=MenCoralLight!40] at (\cueXR,\cueYbot)
  {other person\\[-1pt]\textcolor{MenCoral!85!black}{\tiny$\rightarrow$ wrong person}};
\fill[MenCoralLight!55, rounded corners=2pt] (\xA+0.15,0.55) rectangle (\xA+\colW-0.15,1.35);
\node[font=\scriptsize\bfseries, text=MenCoral!90!black, anchor=west] at (\xA+0.25,1.18) {Passive threshold};
\node[font=\scriptsize, text=MenSlate, anchor=west] at (\xA+0.25,0.82) {speak $\Leftrightarrow e_t>\theta$};

\fill[MenTealLight!30, panel] (\xB,\yBot) rectangle (\xB+\colW,\yTop);
\draw[MenTeal!50, panel] (\xB,\yBot) rectangle (\xB+\colW,\yTop);
\fill[MenTeal!85, rounded corners=2pt] (\xB+0.1,\yHdrBot) rectangle (\xB+\colW-0.1,\yTop-0.06);
\node[hdr, anchor=west] at (\xB+0.2,\yTop-0.22) {\textsc{pas}};
\node[mod, draw=MenTeal!65, fill=white] (m1) at (\pasX,\pasYmid+\pasGap)
  {\textbf{P:\ Probe}\\no speech yet};
\node[mod, draw=MenSlate!40, fill=MenSlateLight!50] (m2) at (\pasX,\pasYmid)
  {\textbf{A:\ Authorize}\\gate on evidence};
\node[mod, draw=MenGold!75!black, fill=MenGold!6] (m3) at (\pasX,\pasYmid-\pasGap)
  {\textbf{S:\ Speak}\\may greet first};
\draw[arr, MenTeal!70] ([yshift=-1pt]m1.south) -- ([yshift=1pt]m2.north);
\draw[arr, MenTeal!70] ([yshift=-1pt]m2.south) -- ([yshift=1pt]m3.north);

\fill[white, panel] (\xC,\yBot) rectangle (\xC+\colW,\yTop);
\draw[MenGold!55, panel] (\xC,\yBot) rectangle (\xC+\colW,\yTop);
\fill[MenGold!80!black, rounded corners=2pt] (\xC+0.1,\yHdrBot) rectangle (\xC+\colW-0.1,\yTop-0.06);
\node[hdr, anchor=west] at (\xC+0.2,\yTop-0.22) {Outcomes};
\fill[MenCoralLight!40, rounded corners=2pt] (\xC+0.15,1.95) rectangle (\xC+\colW-0.15,3.05);
\node[font=\scriptsize\bfseries, text=MenCoral!90!black, anchor=west] at (\xC+0.25,2.88) {Passive};
\node[font=\scriptsize] at (\xC+1.78,2.45) {``Hi!'' $\times$ awkward};
\fill[MenTealLight!45, rounded corners=2pt] (\xC+0.15,0.55) rectangle (\xC+\colW-0.15,1.85);
\node[font=\scriptsize\bfseries, text=MenTeal!90!black, anchor=west] at (\xC+0.25,1.68) {Authorized};
\node[font=\scriptsize] at (\xC+1.78,1.15) {probe $\rightarrow$ ``May I help?'' $\checkmark$};

\draw[arr, MenSlate!50] (\xA+\colW+0.02,\pasYmid+\pasGap) -- (\xB-0.02,\pasYmid+\pasGap)
  node[midway, above=2pt, font=\footnotesize, text=MenSlate] {evidence};
\draw[arr, MenTeal!65] (\xB+\colW+0.02,\pasYmid-\pasGap) -- (\xC-0.02,\pasYmid-\pasGap)
  node[midway, above=2pt, font=\footnotesize, text=MenTeal] {policy};

\fill[MenSlateLight, rounded corners=2pt, draw=MenSlate!18, line width=0.5pt]
  (0.05,-0.08) rectangle (11.0,0.30);
\node[font=\scriptsize, align=center, inner sep=0pt] at (5.52,0.11) {%
  \textbf{Generalist safety} needs context and user preference; harm often starts at the \textbf{first hard-to-undo act}.%
};
\end{tikzpicture}%
}
\caption{Initiation authorization under uncertainty (left to right).
\textbf{User cues} mix what the robot sees with what that should mean (on the
phone so don't speak, a brief glance so the robot is not sure yet, eye contact so
probing is OK, or attention on someone else so the robot has the wrong
person), while many deployed stacks still use a \emph{passive threshold}
(speak $\Leftrightarrow e_t>\theta$) that treats an engagement score
as permission to speak.
\textbf{\textsc{pas}} (probe--authorize--speak) uses small reversible
\emph{probes} that never emit speech, gathers response
\emph{evidence}, and opens an authorization \emph{gate} as the only way to
speak first.
\textbf{Outcomes} contrast a passive early greeting (``Hi!'' $\rightarrow$
awkward) with an allowed sequence (probe $\rightarrow$ ``May I help?''
$\rightarrow$ acceptable).}
\label{fig:initiation}
\end{figure*}

Most deployed stacks treat initiation as a fixed threshold:
speak $\Leftrightarrow e_t > \theta$
\citep{bohus2014}. That treats an engagement score, or a single VLA
rollout, as permission to act, with no context, no user preference,
and no evidence from low-cost probes~\citep{skantze2014speech,bohus2014}.
Figure~\ref{fig:initiation} contrasts this passive threshold with
\textsc{pas} (probe--authorize--speak): small reversible probes
before any hard-to-undo action, an explicit bold/conservative setting,
and runtime checks before foundation-model dialogue.

\Needspace{11\baselineskip}
\noindent Building on this distinction, we make three contributions:
\begin{enumerate}
\item We introduce \emph{initiation authorization} as a distinct safety
layer, separate from physical shields, post-plan VLA guardrails,
post-dialogue preference alignment, and engagement optimization.
\item We illustrate \textsc{pas} (probe--authorize--speak) on a physical
robot and propose an evaluation protocol that would compare three
initiation policies.
\item We elaborate on future work and pose open questions on metrics,
governance, and where initiation ends and foundation-model generation
begins.
\end{enumerate}

\section{Initiation Authorization in Generalist-Robot Safety}
\label{sec:taxonomy}

\subsection{Where Initiation Fits}
\label{sec:placement}

Physical safety (collisions, forces, joint limits) remains
necessary \citep{akalin2021,zhang2025safevla}. HRI already separates
physical safety from how safe people feel
\citep{akalin2021}, and social-error taxonomies list what goes wrong once
interaction is already underway \citep{tian2021socialerrors}. Less covered
is whether the robot should start in the first place
\citep{shi2011spatial,edirisinghe2024shopworker}. We call that
\emph{initiation authorization}. Table~\ref{tab:taxonomy} summarizes
the three safety layers for generalist robots.

\begin{table}[H]
\footnotesize
\centering
\caption{Safety layers for generalist robots.}
\label{tab:taxonomy}
\begin{tabular*}{\columnwidth}{@{\extracolsep{\fill}}>{\raggedright\arraybackslash}p{0.21\columnwidth}>{\raggedright\arraybackslash}p{0.37\columnwidth}>{\raggedright\arraybackslash}p{0.26\columnwidth}@{}}
\toprule
\textbf{Layer} & \textbf{What it protects} & \textbf{Example problem} \\
\midrule
Physical & Contact force, clearance & Unintended touch \\
\midrule
Social-interaction &
  Felt safety, social norms during ongoing contact &
  Interrupt, pressure \\
\midrule
Initiation authorization &
  When to act, whom to address &
  Acting too soon \\
\bottomrule
\end{tabular*}
\end{table}

Initiation is not only about speech. On talking robots the first
hard-to-undo action is often a word. On manipulation VLAs it can be
the first grasp, stepping into someone's space, or a question that
cuts into a human conversation \citep{openvla-2024}.
What matters is whether the action is hard to take back while the
situation is still unclear, not whether it is speech, reach, or
approach. In Table~\ref{tab:themes}, we map common safety concerns to
these before-the-first-act decisions.

\begin{table}[H]
\small
\centering
\caption{Safety concerns mapped to initiation authorization.}
\label{tab:themes}
\begin{tabular*}{\columnwidth}{@{\extracolsep{\fill}}>{\raggedright\arraybackslash}p{0.34\columnwidth}>{\raggedright\arraybackslash}p{0.60\columnwidth}@{}}
\toprule
\textbf{Concern} & \textbf{Question before the first act} \\
\midrule
Venue and context & Same greeting or reach OK in demo, rude at home \\
\midrule
Willingness to interact & Looking at the robot $\neq$ wanting to be greeted \\
\midrule
Social risk preference & Bold greeting vs.\ user who wants privacy \\
\midrule
Speaking too early vs.\ too late & Awkward interrupt vs.\ user not greeted \\
\midrule
Speak, act, and log permissions & Who may speak, move, log, or learn from data \\
\bottomrule
\end{tabular*}
\end{table}

\FloatBarrier

\subsection{Generalist Safety and Related Approaches}

Single-purpose robots are often built on assumptions that generalists
cannot use: (i) one engagement threshold $\theta$ works for everyone
\citep{bohus2014,rich2010}, (ii) observable cues are enough to infer
intent, without reversible probes
\citep{skantze2014speech,broek2024proactive}, (iii) speaking too soon
and waiting too long are equally bad mistakes
\citep{shi2011spatial}, and (iv) the same initiation script works in
every venue \citep{edirisinghe2024shopworker,akalin2021}. Fixed-task
robots mostly worry about collisions and excess force, calibrated once
at deploy. Generalist robots also face social interrupts, uncertain
consent, and settings that call for different bold or conservative
initiation. They need \emph{tunable authorization}: a per-venue and
per-user setting for how bold or conservative to be, without retraining
the whole VLA stack~\citep{harland2024adaptivealignment}.

When these assumptions fail, one safety layer is not enough. A
full stack needs three checks. Physical shields and motion planning
limit contact risk. VLA alignment
\citep{zhang2025safevla} checks whether a chosen plan
is safe once the robot is already trying to act. \emph{Initiation
authorization} runs before the first hard-to-undo social action.
Recent post-plan guardrails \citep{ravichandran2025roboguard} catch
unsafe or jailbroken plans, but only at that later stage. They do not
decide whether the robot should speak, reach, or approach in the first
place \citep{robey2024jailbreaking}. Initiation authorization complements
this nearby work rather than replacing it. Post-plan guardrails and
physical shields ask whether a chosen motion is safe once execution has
started \citep{zhang2025safevla,ravichandran2025roboguard}. Preference
alignment and dialogue safety tools largely apply after interaction has
begun~\citep{bai2022constitutional,robey2024jailbreaking}. Proactive
HRI and engagement scoring focus on how well to initiate, not whether to
initiate when cues are still ambiguous
\citep{broek2024proactive,shi2011spatial,rich2010}. A collision-free,
well-aligned, high-engagement policy can still open with the wrong first
word.

\section{PAS and Evaluation Protocol}
\label{sec:pas}
We implement initiation authorization as \textsc{pas}
(probe--authorize--speak) on PAL Robotics' ARI humanoid robot. \textsc{pas}
combines an engagement score, staged probe rules, and a threshold gate
\citep{bohus2014,skantze2014speech,rich2010} as a safety layer with (i) a probe module structurally prevented from
emitting speech, (ii) a gate threshold $\tau(\rho)$ that is a
deploy-time dial rather than a fixed trained constant, and (iii) a
logged margin $\Delta_{\text{init}}$ at the moment the robot says its
first word to someone.

\subsection{System}

The controller runs at 10\,Hz. Each cycle follows three steps:
perception, staged probes, then the authorization gate. Only the gate
can trigger the robot's first word. The dialogue model runs only after
that.

\textbf{Step 1: Perception.} The system tracks visible people and fuses
gaze, speech activity, approach, and dwell into an engagement score $e_t$.

\textbf{Step 2: Staged probes.} The robot moves through four levels of
nonverbal signaling. At each level it may only \emph{wait}, \emph{turn its
head toward the person}, or \emph{turn its body to face the person}, and
never speak. A higher level requires stronger signs that the person is open to
interaction, and the level can drop back if the person looks away.
Table~\ref{tab:probe_stages} shows the person cues and robot actions
at each level.

\begin{table}[H]
\footnotesize
\centering
\caption{Four probe levels: person cues and robot actions.}
\label{tab:probe_stages}
\begin{tabular*}{\columnwidth}{@{\extracolsep{\fill}}c>{\raggedright\arraybackslash}p{0.36\columnwidth}>{\raggedright\arraybackslash}p{0.44\columnwidth}}
\toprule
\textbf{Level} & \textbf{Person must show} & \textbf{Robot does} \\
\midrule
0 & Someone nearby & Stays still \\
1 & Brief eye contact & Turns head toward person \\
2 & Sustained attention, not walking away & Turns body to face person \\
3 & Stays nearby several seconds & Keeps facing person, gate may open \\
\bottomrule
\end{tabular*}
\end{table}

\textbf{Step 3: Authorization gate.} The gate releases the first word
only when $e_t \geq \tau(\rho)$ and probe level $\geq 2$, where $\tau(\rho)$
is set by the venue bold/conservative dial $\rho$
\citep{harland2024adaptivealignment}. The system logs
$\Delta_{\text{init}} = e_t - \tau(\rho)$ at that moment.

\begin{algorithm}[h]
\caption{PAS initiation loop (10\,Hz)}
\small
\begin{algorithmic}[1]
\State \textbf{Input:} frame, stage, $\rho$
\State $e_t \leftarrow \text{engagement\_score}()$
\State $\tau \leftarrow \tau(\rho)$
\State $a_t \leftarrow \text{staged\_probe}(e_t,\text{stage})$
\Comment{never speak}
\If{$e_t \geq \tau$ \textbf{and} stage $\geq 2$}
  \State $a_t \leftarrow$ speak \Comment{first word to user}
  \State log $\Delta_{\text{init}} = e_t - \tau$
  \State run dialogue model
\EndIf
\end{algorithmic}
\end{algorithm}

\begin{figure}[H]
\centering
\includegraphics[width=\columnwidth]{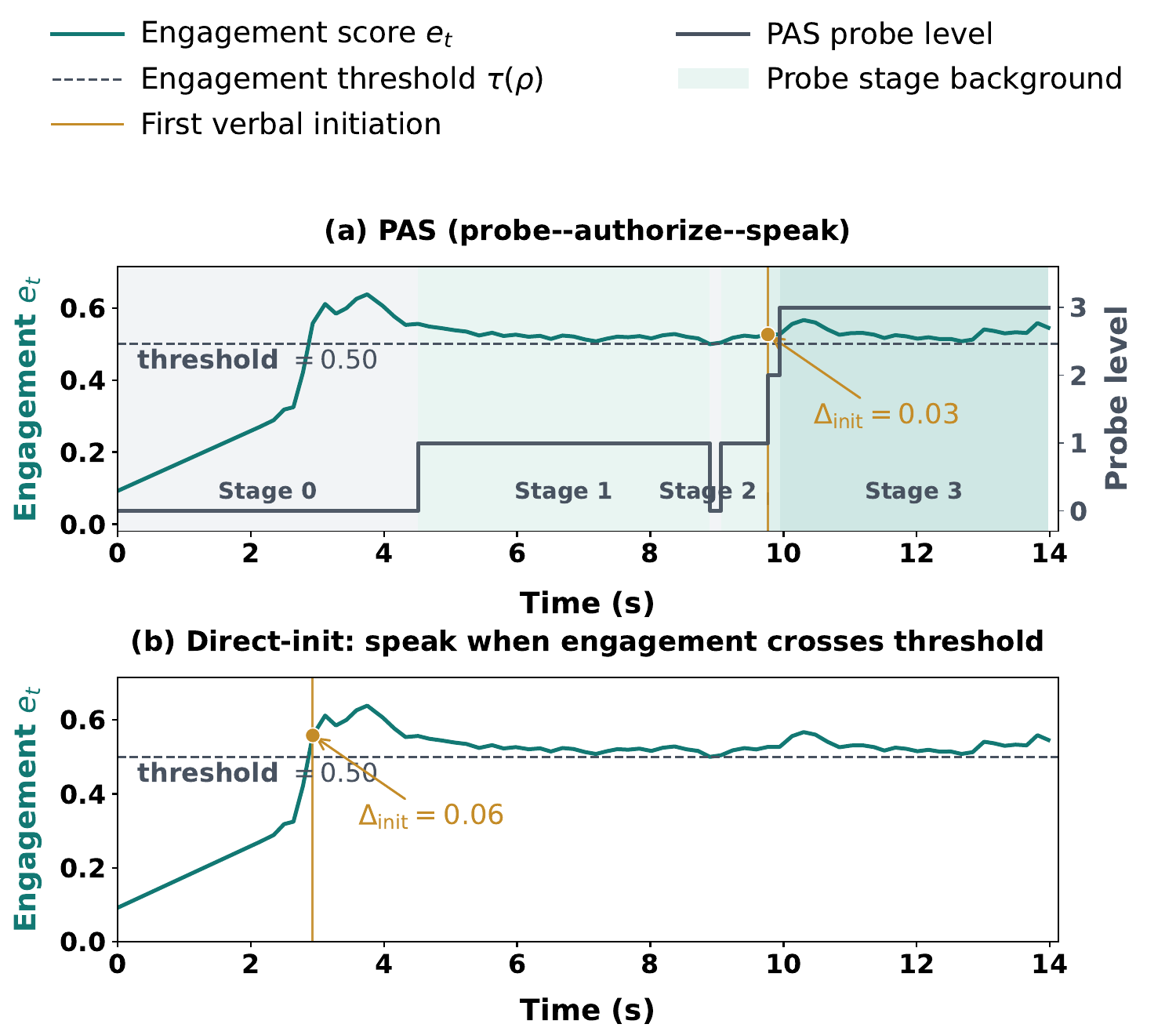}
\caption{Replay of one logged $e_t$ trace under two initiation policies.
\mbox{\textbf{(a)~PAS:}} staged probes, then gate.
The first word fires at $t{\approx}10$\,s only once $e_t{\geq}\tau(\rho)$
and probe level ${\geq}2$ are both satisfied
($\Delta_{\text{init}}{=}0.03$).
\mbox{\textbf{(b)~Direct-init:}} threshold only.
The first word fires at $t{\approx}3$\,s once $e_t{\geq}\theta$
($\Delta_{\text{init}}{=}0.06$).
Although $e_t$ crosses the threshold at $t{\approx}3$\,s, PAS withholds
speech until staged evidence catches up.
The delay reflects structural authorization rather than a change in the
engagement signal.}
\label{fig:pas_trace}
\label{fig:trace}
\end{figure}

Figure~\ref{fig:pas_trace} replays one logged doorway trace under
\textsc{pas} and direct-init, showing staged probes, gate crossing, and
$\Delta_{\text{init}}$ at the first word.

\subsection{Proposed Evaluation}

We propose a between-subjects study in which each participant completes
one doorway session under one initiation policy, while the robot,
perception stack, scenario, and dialogue stack stay fixed
(Table~\ref{tab:conditions}).

\begin{table}[H]
\small
\centering
\caption{Three initiation policies in the proposed study.}
\label{tab:conditions}
\begin{tabular*}{\columnwidth}{@{\extracolsep{\fill}}>{\raggedright\arraybackslash}p{0.28\columnwidth}>{\raggedright\arraybackslash}p{0.66\columnwidth}@{}}
\toprule
\textbf{Condition} & \textbf{What it does} \\
\midrule
\textsc{pas} & Probes + gate (default $\rho$) \\
\midrule
Direct-init & $e_t > \theta$, no probes or gate \\
\midrule
Passive-wait & Probes + gate with conservative $\rho$ \\
\bottomrule
\end{tabular*}
\end{table}

\paragraph{Scenario.} Sessions would start at a scripted doorway. Each
participant would follow one script: on a phone, mid-conversation, or
clearly open to interaction. Reach, approach, and multi-party
authorization are deferred to \S\ref{sec:futurework}.

\paragraph{Endpoints.} Sessions would log probe actions, gate state, probe
stage, and $\Delta_{\text{init}}$ at the robot's first word, plus
user-reported success, awkwardness, and naturalness on 7-point Likert
scales. Subjective ratings would compare all three policies. Logged
$\Delta_{\text{init}}$ would compare \textsc{pas} with direct-init
(Fig.~\ref{fig:pas_trace}); passive-wait uses the same probe--gate stack
with a conservative $\rho$. On logged sessions we would also vary $\rho$
to see how a bolder or more conservative dial changes how often the
robot speaks and how large $\Delta_{\text{init}}$ is at the first word.

\section{Discussion}
\label{sec:discussion}
\subsection{Future Work}
\label{sec:futurework}

After the doorway study, we plan three extensions. First, extend \textsc{pas} from
single-encounter doorways to multi-party scenes, where choosing whom
to speak to becomes part of authorization. Second, integrate
\textsc{pas} with VLA action policies, so that initiation authorization
gates not only speech but reach and approach. Third, release
the initiation log format as an open schema, supporting cross-platform
comparison of first-act decisions.

\subsection{Open Questions}

Retail, home, and museum settings differ but share one pattern: the
robot acts before knowing who is willing to be addressed
\citep{edirisinghe2024shopworker,shi2011spatial}. This raises three
families of open questions.

\paragraph{Measurement and standards.} We do not yet agree on how to
score whether a robot spoke or acted too soon. Should safety checklists
list that problem on its own, instead of folding it into engagement
scores? What should every study at least report: how awkward people
felt, whether tasks were interrupted, whether the robot picked the
right person, and the margin at the first word (such as
$\Delta_{\text{init}}$)? Without numbers on that first act, a robot can
look safe overall and still greet someone on a phone too often in the
doorway cases that matter most \citep{akalin2021}.

\paragraph{Preference and governance.} Who should set how bold or shy
the robot is: the deployer, the user, the platform, or a regulator?
Can one policy fit every generalist robot, or do homes, shops, and
clinics each need their own safety case? A home user may want the robot
to offer help early; a clinic may need it to wait. Operators need a
dial they can read and change after an incident, not a hidden threshold
buried in model weights.

\paragraph{Foundation models and layering.} Dialogue models may sound
more natural without knowing better \emph{when} to start
\citep{ravichandran2025roboguard}. Where should ``what to say'' stop
and a runtime check on ``may I start now?'' begin? Some failures show up
only when those two jobs are merged: greeting the wrong person, a
jailbroken opening line, or talking too much too soon. Safety tests
should include the first physical act. Adversarial prompts can trigger
a bad first move, not just a toxic reply later, and text-only benchmarks
will not catch that \citep{robey2024jailbreaking}. End-to-end VLAs
\citep{openvla-2024} raise the stakes because one rollout can combine
seeing, deciding, and acting in a single step.

\section{Conclusion}
\label{sec:conclusion}
Before motion or dialogue safety, generalist robots need one more
check: should they take a first hard-to-undo social action now? We call
that \emph{initiation authorization}. The harm is often a mistimed
greeting or reach (for example, to someone on a phone), not a
collision; a robot can be collision-free and still feel unsafe at first
contact. We implement this in \textsc{pas} with reversible probes, an
authorization gate, and only then dialogue generation. A ``safe
generalist robot'' must mean more than collision-free motion or aligned
dialogue; it must also mean that the robot knows when not to start.

\balance
\bibliographystyle{plainnat}
\bibliography{references}

\end{document}